%% file: _main.tex
\input{_constants}
\arxiv 

\pdfoutput=1
\documentclass[10pt,twocolumn,letterpaper]{article}
\usepackage{booktabs}
\usepackage{color}
\input{cvpr_header}
\begin{document}
\title{Local Contrast and Global Contextual Information Make Infrared Small Object Salient Again}
\author{\authorBlock}
\maketitle

\input{00_abstract}
\input{01_intro}
\input{02_related}
\input{03_method}
\input{04_experiments}
\input{10_conclusion}

{\small
\bibliographystyle{ieee_fullname}
\bibliography{11_references}
}

\ifarxiv \clearpage \input{12_appendix} \fi

\end{document}

%% file: _constants.tex
\def\cvprPaperID{5850}
\def\confName{ICCV}
\def\confYear{2023}

\def\authorBlock{
    Chenyi Wang \qquad
    Huan Wang \qquad
    Peiwen Pan \\
    NJUST \\
    {\tt\small \{wcyjerry, Nanjing\}@qq.com}
}

\newif\ifreview 
\newif\ifarxiv \newcommand{\arxiv}{\arxivtrue}
\newif\ifcamera 
\newif\ifrebuttal 

%% file: cvpr_header.tex
\ifreview \usepackage[review]{cvpr} \fi
\ifarxiv \usepackage[pagenumbers]{cvpr} \fi
\ifrebuttal \usepackage[rebuttal]{cvpr} \fi
\ifcamera \usepackage{cvpr} \fi

\usepackage{graphicx}
\usepackage{amsmath}
\usepackage{amssymb}
\usepackage{booktabs}

\input{_macros}  

\usepackage{xr-hyper}

\makeatletter
\newcommand*{\addFileDependency}[1]{
  \typeout{(#1)}
  \@addtofilelist{#1}
  \IfFileExists{#1}{}{\typeout{No file #1.}}
}

\makeatother

\usepackage[pagebackref,breaklinks,colorlinks]{hyperref}
\usepackage[capitalize]{cleveref}
\crefname{section}{Sec.}{Secs.}
\crefname{table}{Table}{Tables}
\crefname{figure}{Fig.}{Figs.}

%% file: _macros.tex
\usepackage{times}
\usepackage{microtype}
\usepackage{epsfig}
\usepackage[table,xcdraw]{xcolor}
\usepackage{caption}
\usepackage{float}
\usepackage{placeins}
\usepackage{color, colortbl}
\usepackage{stfloats}
\usepackage{enumitem}
\usepackage{tabularx}
\usepackage{xstring}
\usepackage{multirow}
\usepackage{xspace}
\usepackage{url}
\usepackage{subcaption}
\usepackage{xcolor}
\usepackage[hang,flushmargin]{footmisc}

\ifcamera \usepackage[accsupp]{axessibility} \fi

% ...

\ifarxiv  \fi

\newcommand{\R}[1]{{%
    \textbf{%
        \ifstrequal{#1}{1}{\textcolor{red}{R#1}}{%
        \ifstrequal{#1}{2}{\textcolor{blue}{R#1}}{%
        \ifstrequal{#1}{3}{\textcolor{magenta}{R#1}}{%
        \ifstrequal{#1}{4}{\textcolor{teal}{R#1}}{%
                           \textcolor{cyan}{R#1}%
        }}}}%
    }%
}}

%% file: 00_abstract.tex
\begin{abstract}
Infrared small object segmentation (ISOS) aims to segment small objects only covered with several pixels from clutter background in infrared images.
It's of great challenge due to: 1) small objects lack of sufficient intensity, shape and texture information; 2) small objects are easily 
lost in the process where detection models, say deep neural networks, obtain high-level semantic features and image-level receptive fields through successive downsampling.
This paper proposes a reliable segmentation model for ISOS, dubbed UCFNet (U-shape network with central difference convolution and fast Fourier convolution), which can handle well the two issues. It builds upon central difference convolution (CDC) and fast Fourier convolution (FFC). 
On one hand, CDC can effectively guide the network to learn the contrast information between small objects and the background, as the contrast information is very essential in human visual system dealing with the ISOS task. On the other hand, FFC can gain image-level receptive fields and extract global information on high-resolution features maps while preventing small objects from being overwhelmed. 
Experiments on several public datasets demonstrate that our method significantly outperforms the state-of-the-art ISOS models, and can provide useful guidelines for designing better ISOS deep
models. Code are available at https://github.com/wcyjerry/BasicISOS.
\end{abstract}

%% file: 01_intro.tex
\vspace{-5mm}
\section{Introduction}
\label{sec:intro}
Infrared small object segmentation (ISOS) is a key technique broadly used in early warning systems, night navigation, maritime surveillance, UAV search and tracking and the like, due to its all-weather working, long-range detection and concealment characteristics. Therefore, improving its performance is 
of great significance.
\begin{figure}[t]
    \centering
    \includegraphics[width = 1\linewidth]{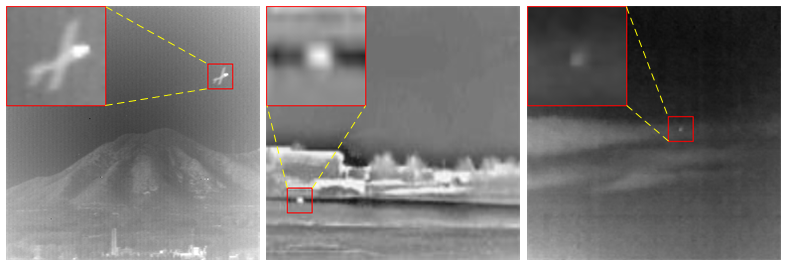}
    \caption{Examples of ISOS with the object indicated by the red bounding boxes and a close-up is shown in the top left corner. Left: a airplane with recognizable shape in a remote distance. Middle: a car whose shape is almost lost and only can be identified by estimation. Right: A small dim object drowned in a cloud background.\vspace{-6mm}}
    \label{first_pig}
\end{figure}
Researches on ISOS have been conducted for over several decades. Many methods are proposed which can be roughly categorized into (1) traditional methods focusing on signal processing and prior knowledge and (2) deep learning 
models relying on Convolutional Neural Networks (CNNs) and Visual Transformers(ViTs).

Traditional methods consist of three representative subcategories: background-oriented methods, object-oriented methods, 
and low-rank decomposition methods. Background-oriented methods like Max-mean/Max-medium\cite{Deshpande1999MaxmeanAM} and Top-Hat\cite{Zeng2006TheDO} separate object from
complex background by using all kinds of filters to estimate the scene background. Object-oriented methods segment small object by designing different measure methods. For instance, LCM\cite{chen2013local}, ILCM\cite{han2014robust} and TLLCM\cite{han2019local} use the contrast measure 
between central point and its surroundings and PatchSim\cite{Bai2016PatchSB} applies patch similarities to suppress false alarms.
Low-rank decomposition methods are frequently based on robust principal component analysis(RPCA), by inductively treating the input as a superposition 
of low-rank background and sparse objects and solving such detection issues via optimization techniques. Take the infrared patch-image (IPI)\cite{Gao2013InfraredPM} model as an example. It suggests a patch-sliding design that exploits better non-local self-correlation properties of images via RPCA. 
Subsequently, recent works put efforts into designing sound low-rank and prior constraints\cite{rawat2020reweighted,wan2021total}, 
exploiting spatio-temporal and multi-mode correlation\cite{dai2017reweighted}, and applying advanced optimization schemes\cite{dai2016infrared}. 
Though traditional methods have achieved some results in experimental scenarios, they are sensitive to hyper-parameter setting, lack of generalization and suffer from low performance under complex real scenes.

As deep learning has become the mainstream in many computer vision tasks, many pioneers have achieved great improvement in ISOS using deep neural networks. 
Wang et al. \cite{wang2019miss} used two generators to focus on two different tasks miss detection and false alarm and one discriminator to get the balanced result of two generators. Dai et al. \cite{dai2021asymmetric} proposed a novel feature fusion method named asymmetric contextual modulation (ACM), and proposed the first 
public ISOS dataset in real scenes. Then they proposed ALC \cite{TGRS21ALCNet} which included an unlearnable conditional local contrast module. Though existing deep learning methods have got great results, they mostly focus on the feature fusion. Liu et al. \cite{liu2021infrared} first introduced transformer block into ISOS task and got great results. Zhang et al. introduced attention-guided context module to help AGPCNet\cite{zhang2021agpcnet} focus on small object. Zhang et al. proposed ISNet\cite{zhang2022isnet} which take shape into account to achieve better performance. 

However, the above models more or less neglect two native problems in ISOS: the infrared small object lacks of sufficient common information like color and shape and small objects would be drown in the excessive downsampling process. This paper aims to provide a better solution to the problems of information lacking and objects being drown. Specifically, we design a U-net network behaving as our ISOS backbone. Then we employ central difference convolution (CDC), a human visual system based convolution operator, to extract essential contrast information. As for vanish problem, we leverage fast Fourier convolution (FFC) which uses convolution operation on frequency domain, thus being easy to obtain global information while avoiding object disappearance. We demonstrate the effectiveness of CDC with other advanced convolution operators and we also compare FFC with other global information extracting methods. Our final model, coined as UCFNet, achieves new state-of-the-art performance over competitive ISOS methods on existing public datasets.
The contributions of our work can be summarized as: 

\begin{itemize}
    \item We reveal two native problems in infrared small object segmentation through analysing experiments with common segmentation methods and propose a baseline network structure. 
    \item We further propose CDC to extract local contrast information based on human visual system which is essential for infrared small object detection, meanwhile we use FFC to obtain image-level receptive fields and global context while maintaining high resolution. 
    \item We comprehensively evaluate our proposed method on two public datasets, our UCF achieves new state-of-the-art performance over other methods.
\end{itemize}

%% file: 02_related.tex
\section{Related Work}
\label{sec:related}
\subsection{ISOS}
There are two main types of ISOS methods, traditional methods based on mathematical modeling and deep learning methods based on neural networks.

Among the traditional methods, Max-mean-Max-medium\cite{Deshpande1999MaxmeanAM} and Top-Hat\cite{Zeng2006TheDO} used filters to separate target of background.
LCM\cite{chen2013local}, ILCM\cite{han2014robust}, TLLCM\cite{han2019local} and MPCM\cite{wei2016multiscale} segmented small object by designing salient measures.
IPI model\cite{Gao2013InfraredPM} treated the input as a superposition of low-rank background and sparse yet shaped target, and solved such issues by using Low-rank decomposition, further methods like sound low-rank\cite{rawat2020reweighted} and prior constraints\cite{wan2021total} were proposed based on IPI.
These methods suffered from low performance under complex scenarios.

As for deep learning methods, Wang et al. \cite{wang2019miss} used conditional GAN\cite{mirza2014conditional} with two generators and one discriminator to gain  
a great balance between miss detection and false alarm (MDvsFA). Dai et al. \cite{dai2021asymmetric} proposed an asymmetric contextual modulation to help network performance
well and introduced the first public ISOS dataset SIRST in real scenes, Dai et al. \cite{TGRS21ALCNet} further applied a handcraft dilated local 
contrast measure into network. Liu et al. \cite{liu2021infrared} firstly introduced multi-head self-attention into ISOS tasks and got a good result. Zhang et al. \cite{zhang2021agpcnet} proposed AGPCNet with attention-guided context block and context pyramid module.
Zhang et al. \cite{zhang2022isnet} took shape into account and designed Taylor finite difference edge block and two-orientation attention, and also proposed a more challengable dataset IRSTD.
Deep learning methods have become more and more dominant in ISOS due to their great ability of robustness and generalization. 

\subsection{Convolution Operators}
The ordinary convolution operation is to divide the feature map into patches of the same size as the convolution kernel and then 
perform a weighted sum operation, this fixed operation at each position may be suboptimal for some specific tasks. Therefore, 
researchers have proposed some advanced convolution operators.
In order to solve the problem of standard convolution treating all input pixels as valid ones in image inpainting tasks,
Liu et al. \cite{liu2018image} proposed partial convolution, where the convolution is masked and renormalized to be conditioned
on only valid pixels. Yu et al. \cite{yu2019free} proposed gated convolution to provide a learnable dynamic feature selection mechanism for each channel at 
each spatial location across all layers.
While in modeling geometric transformations, vanilla convolution is limited due to its fixed geometric structure,
Dai et al. \cite{dai2017deformable} proposed deformable convolution to enhance this ability by adding additional offsets learned from target task.
Zhu et al. \cite{zhu2019deformable} then proposed deformable convolution v2 to reduce the impact of irrelevant region by adding additional 
weight and mimicking features from RCNN\cite{ren2015faster}.
Yu et al. \cite{yu2020searching} proposed central difference convolution in face anti-spoofing task, which is able to capture intrinsic detailed patterns via aggregating both intensity and gradient information.
\subsection{Global Information Extraction}
\label{subsec:global_information_extraction}
One common concern is the ability of the network to grasp the local and global information, local context usually are easy to extract using convolution, while how global the information can get usually 
determined by the receptive fields of the network. The most common strategy to enlarge the receptive fields is stacking convolutions and downsamplings constantly,
continuous convolution can linearly enlarge receptive fields while downsamplings multiply,
dilated (atrous) convolution inserts holes between pixels in convolutional kernels and can obtain larger receptive fields than standard convolution, Chen et al. \cite{chen2017deeplab} proposed dilation atrous pyramid 
pooling (ASPP)
to capture multi-scale information and Wang et al. \cite{wang2018understanding} designed a hybrid dilated convolution
to enlarge receptive fields, they both got good improvements in semantic segmentation tasks.
Attention mechanism \cite{wang2018non,vaswani2017attention} can gain global information by calculating the correlation of each single pixel 
with other pixels. Fu et al. proposed spatial attention and channel attention \cite{fu2019dual} and achieved great results.
Chi et al. \cite{chi2020fast} proposed fast Fourier convolution which performs convolution operators in frequency domain to conduct a global influence in spatial domain thus it  can extract image-level information. Suvorov et al. proposed LAMA\cite{suvorov2022resolution} which successfully applied it in large kernel image inpainting tasks. Berenguel et al. \cite{berenguel2022fredsnet} then used FFC in monocular depth estimation and semantic segmentation.

%% file: 03_method.tex
\section{Method}
\label{sec:method}
\subsection{Inspiration}
\label{subsec:Preliminary}
\begin{figure}
    \centering
    \includegraphics[width=1\linewidth]{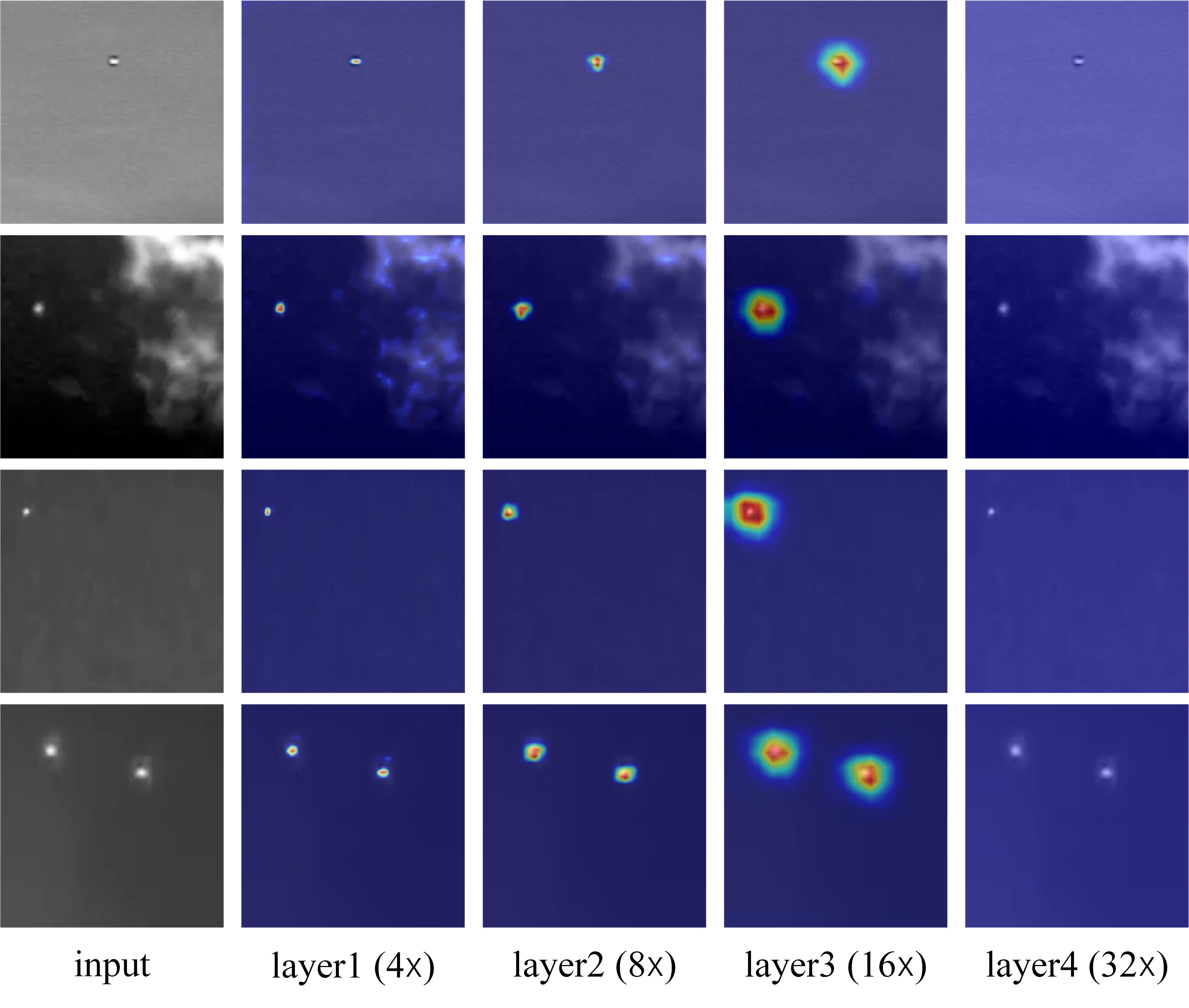}
    \caption{Visualization of 4 layers in FPN.  Objects are completely lost in layer 4 with a 32$\times$ downsampling.}
    \label{grad_cam}
\end{figure}

We draw inspiration from a series of  experiments using common segmentation networks, including FPN\cite{lin2017feature}, U-Net\cite{ronneberger2015u}, PSPNet\cite{zhao2017pyramid} 
and DeepLabv3\cite{chen2017deeplab}, the results in \cref{common models results table} indicate two anomalies: first, the performance is quite polarized, with the models that fuse more low-level features (FPN and U-Net) significantly outperforming the others; second, their performance drops only slightly as the network width and depth increase. 
\begin{table}[thb]
    \centering
    \caption{Results of common segmentation models in ISOS tasks.}
    \renewcommand{\arraystretch}{1.4}
    \scalebox{0.75}{
    \begin{tabular}{ l l c c c }
    \hline
        \textbf{Method}&\textbf{Backbone}&\textbf{Base width}&\textbf{IoU}$\uparrow$&\textbf{Params (M)$\downarrow$}\\
    \hline
    U-Net\cite{ronneberger2015u}&-&64&69.89&31.04\\
    \hline
    \multirow{3}{*}{FPN\cite{lin2017feature}}
    &ResNet-18&64&70.38&14.12  \\
    &ResNet-50&256&70.12&27.19 \\
    &ResNet-101&256&69.64&46.18 \\
    \hline
    PSPNet\cite{zhao2017pyramid}    
    &ResNet-50&256&22.73&46.65
    \\
    \hline
    DeepLabv3\cite{chen2017deeplab}&ResNet-50&256&39.71&35.86\\
    \hline
    \end{tabular}
    }
    \label{common models results table}
\end{table}
We further visualize the attention maps of 4 stages using grad-cam\cite{selvaraju2017grad,jacobgilpytorchcam} in FPN. As we can see in \cref{grad_cam}, the small object has been completely lost due to the excessive downsampling. Meanwhile directly increasing the width and depth of network still fails to mine more information while may lead to redundancy of parameters. Based on these observation, we summarize two essential guidelines on ISOS: 1) Extracting global information while maintaining high resolution may avoid small object being overwhelmed. 2) Extracting additional essential information using modules dedicated to ISOS could be more effective than directly increasing the depth and width of a network.

\subsection{Overview of UFCNet}
\label{subsec:Overview of UFCNet}
\begin{figure*}[tp]
    \centering
    \includegraphics[width=1\linewidth]{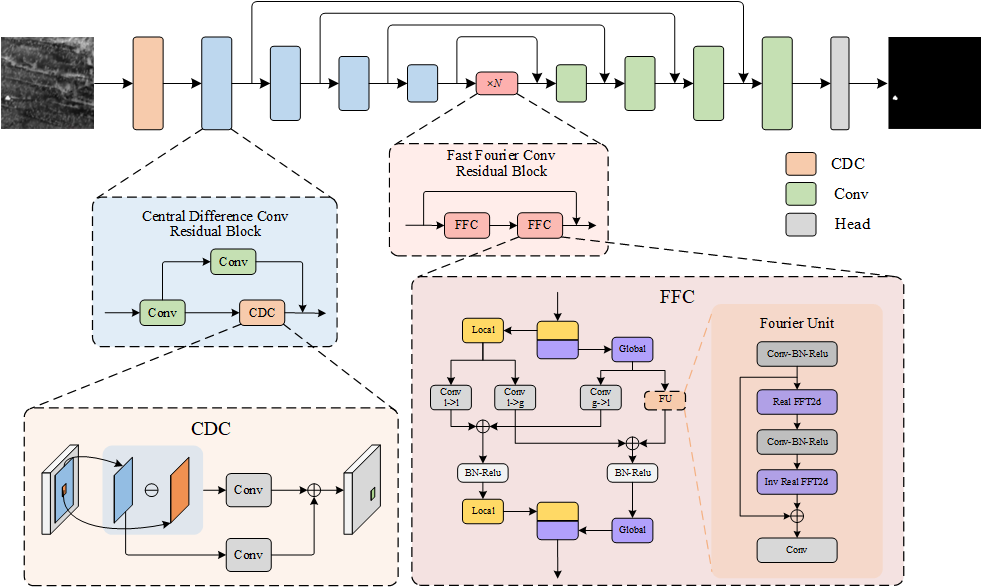}
    \caption{The scheme of the proposed model UCFNet. UCFNet is based on central difference convolution residual blocks and fast Fourier convolution residual blocks, can effectively extract both local contrast information and global context.}
    \label{UCFNet}
\end{figure*}

Based on the enlightenment from \cref{subsec:Preliminary}, we proposed a U-shape network with central difference convolution (CDC) and 
fast Fourier convolution (FFC). The whole framework can been seen in \cref{UCFNet}. The number of base channels is 32 and only 4 downsampling operations
are performed in the whole process, which satisfies the needs of ISOS tasks while avoiding inefficient redundancy. Specially, during the downsample process, we use central difference convolution residual block which contains a standard convolution and CDC with a residual convolution for channel alignment. We use two cascaded FFC to form our fast Fourier convolution residual block which aims to extract global context in high resolution. In what follows, 
we will briefly introduce central difference convolution in \cref{subsec:cdc} and fast Fourier convolution in \cref{subsec:ffc}.

\subsection{Central Difference Convolution}
\label{subsec:cdc}
Convolution can effectively extract color, shape, texture and other information, while standard convolution may be limited because the lack of
these information in ISOS tasks. Inspired by human visual system sensitive to intensity difference and contrast, we utilize central difference convolution to guide our network.
CDC can introduce additional contrast information by computing the difference between the center point and other points within the convolution 
window and can be described as \cref{CDC},
\begin{align}
    \label{CDC}
    f(x,y) = \sum_{(i,j)\in R}w_{(i,j)}\cdot(F_{(x+i,y+j)}-F_{(x,y)})
\end{align}
In addition, CDC is often combined with vanilla convolution to retain some of the traditional feature extraction capabilities, and the whole process can be expressed in the equation.
\begin{align}
    \label{CDC_CONV}
    CDC(x,y) =& \theta \cdot \sum_{(i,j)\in R }w_{(i,j)}(F_{(x+i,y+j)}-F_{(x,y)}) \notag \\
    &+(1-\theta)\cdot \sum_{(i,j)\in R }w_{(i,j)}(F_{(x+i,y+j)})
\end{align}
where the hyperparameter $\theta \in (0,1)$ is used to determine the contribution between CDC and vanilla convolution. The window size $R$ used to calculate the difference is
equal to the convolution kernel size while the receptive fields of it can naturally increase during the forward process, thus making it capable of extracting multi-level contrast information which further help the network identify infrared small objects with different size and effectively guide the network performs well.

\subsection{Fast Fourier Convolution}
\label{subsec:ffc}
Another issue of ISOS is that the small objects are easily overwhelmed in the excessive downsample layers which is used to obtain global information. In order to solve this contradiction, rather than using the Transformer model with long range dependency, we employ fast Fourier convolution (FFC) which can gain image-level receptive fields in high resolution thus we can extract global context without losing small objects. We follow the structure in LAMA\cite{suvorov2022resolution} for FFC. Specifically, we splits the channels into two parallel branch, the local branch uses standard convolution to extract local information while global branch applies the Fourier Unit(FU) to gain global context, and there are two additional short-cuts for information fusion.
The whole process can described as:
\begin{align}
    Y_l = Conv_{l->l}(x_l) + Conv_{g->l}(x_l)\\
    Y_g = Conv_{l->g}(x_g) + FU_{g->g}(x_g)
\end{align}
The Four Unit is key to get global context, because it transforms input features from spatial domain to frequency domain where each single point corresponds 
to all points in the spatial domain. Therefore, the convolution operation being conducted within a small kernel size is able to influence the all image in spatial domain and obtain global contextual information. The FU makes following steps:
\begin{enumerate}
    \item[1)]Transform the input feature map from spatial domain to frequency domain with Real \text{FFT2d} and concatenates the real and imaginary parts
        \begin{align}
        \text{Real}\  \text{FFT2d}&:\mathbb{R}^{H\times W\times C} \rightarrow \mathbb{C}^{H\times \frac{W}{2}\times C} \notag\\
    \text{ComplexToReal} &: \mathbb{C}^{H\times \frac{W}{2}\times C} \rightarrow
        \mathbb{R}^{H\times \frac{W}{2}\times 2C} \notag
        \end{align}
    \item[2)] Apply convolution, normalization and activation function in the frequency domain
    \begin{equation}
        Conv \circ Norm \circ Act: \mathbb{R}^{H\times \frac{W}{2}\times 2C} \rightarrow \mathbb{R}^{H\times \frac{W}{2}\times 2C} \notag
    \end{equation}
    \item[3)] Transform inversely from frequency domain to spatial domain
        \begin{align}
        \text{RealToComplex}&: \mathbb{R}^{H\times \frac{W}{2}\times 2C} \rightarrow
        \mathbb{C}^{H\times \frac{W}{2}\times C} \notag \\
        \text{Inverse Real}\ \text{FFT2d}&: \mathbb{C}^{H\times \frac{W}{2}\times C} \rightarrow \mathbb{R}^{H\times W\times C}\notag
        \end{align}
\end{enumerate}
We concatenate the local branch and global branch into one in the end, and it contains rich local and global information.

%% file: 04_experiments.tex
\section{Experiment}
\label{sec:experiment}
\subsection{Datasets and Metrics}
\label{subsec:datasets and metrics}
\textbf{Datasets.} We evaluate our methods on two widely-used datasets in ISOS:
SIRST\cite{dai2021asymmetric} and IRSTD\cite{zhang2022isnet}. SIRST contains 427 images in real IR scenes with half of the objects in SIRST only contains $0.1\%$ pixels of whole image. Larger than SIRST, IRSTD contains 1001 images with more challenging object and complex backgrounds. Both SIRST and IRSTD are separated into training set and test set with a ratio of 8:2.

\textbf{Metrics.} Following privious works, we use pixel-level metrics (IoU and nIoU) and target-level metrics (Pd and Fa) 
to measure our method. Intersection over Union (IoU) and normalized Intersection over Union (nIoU) can described as:
\begin{align}
    IoU&=\frac{1}{n}\cdot \frac{\sum_{i=0}^{n}tp_i}{\sum_{i=0}^{n}(fp_i+fn_i-tp_i)}\\
    nIoU&=\frac{1}{n}\cdot \sum_{i=0}^{n}\frac{tp_i}{fp_i+fn_i-tp_i}
\end{align}
Where $n$ means to the total number of samples, $tp$ denotes the true positive, $fp$ denotes false positive and $fn$ denotes false negative. While target-level metrics probability of detection (Pd) and false alarm rate (Fa) can be described as:
\begin{align}
	Pd&=\frac{1}{n}\cdot \sum_{i=0}^{n}\frac{N_{pred}^i}{N_{all}^i}\\
	Fa&=\frac{1}{n}\cdot \sum_{i=0}^{n}\frac{P_{false}^i}{P_{all}^i}
\end{align} 
where $N_{pred}$, $N_{all}$ denotes the number of correct detected objects and the number of total objects and $P_{false}$, $P_{all}$ denotes the pixels of false detected objects and the pixels of total objects. We regard that the detection is correct when the distance between the centers of the predict result and the ground truth is less than 4. 

\subsection{Implementation Details}
\label{implementation details}
 We conduct experiments on a computer with a 2.50GHz CPU, 16GB RAM and GeForce RTX 3090 based on Pytorch. For more details, we use AdamW optimizer with an initial learning rate of 0.001 and decayed by Cosine-Anneling-LR\cite{loshchilov2016sgdr} schduler and we use binary cross entropy loss and soft IoU loss as our criterion. Each experiment is trained for 300 epochs with a batch size of 8. Our UCF achieves the best performance with a CDC ratio $\theta$ of 0.7 while using 7 FFC residual blocks.

\subsection{Quantitative Results}
\label{quantitative results}
We evaluated several traditional, deep learning for ISOS and common segmentation methods and compared their results, which are shown in \cref{total results}. Overall, deep learning methods outperformed traditional ones due to their powerful feature extraction and generalization capabilities. Our proposed method (UCF) demonstrated superior performance over other deep learning methods on both datasets, as evidenced by all metrics.
Notably, on SIRST, UCF achieved an impressive performance of 80.89 IoU and 78.72 nIoU, representing an great improvement over other methods, while also achieving a perfect detection rate and an extremely low false alarm rate of only $2.22\times10^{-6}$. On IRSTD, our method also outperformed state-of-the-art deep learning methods by 3-8 points in IoU and nIoU, achieving scores of 68.92 and 69.26, respectively.


\begin{table*}[tp]
    \centering
    \renewcommand\arraystretch{1.3}{
    \setlength{\tabcolsep}{11pt}{

    \begin{tabular}{llllllllll}
\hline
          &                                      & \multicolumn{4}{c}{SIRST}                                          & \multicolumn{4}{c}{IRSTD}                                          \\ \cmidrule(lr){3-6} \cmidrule(lr){7-10}
          &                                      & \multicolumn{2}{l}{Pixel Level} & \multicolumn{2}{l}{Object Level} & \multicolumn{2}{l}{Pixel Level} & \multicolumn{2}{l}{Object Level} \\     \cmidrule(lr){3-4} \cmidrule(lr){5-6} \cmidrule(lr){7-8}
    \cmidrule(lr){9-10}
Method    &  & IoU            & nIoU           & Pd             & Fa              & IoU            & nIoU           & Pd             & Fa              \\ \cmidrule{1-10}
Top-Hat\cite{Zeng2006TheDO}   &   \multirow{8}{*}{\rotatebox{90}{Traditional methods}}                                   & 5.86           & 25.42          & 78.90          & 1397.12         & 4.26           & 15.08          & 67.00          & 422.25          \\
LCM\cite{chen2013local}       &                                      & 6.84           & 8.96           & 77.06          & 183.15          & 4.45           & 4.73           & 57.58          & 66.56           \\
WLDM\cite{WLCM}      &                                      & 22.28          & 28.62          & 87.16          & 98.34           & 9.77           & 16.07          & 63.97          & 177.35          \\
NARM\cite{NARM}      &                                      & 25.95          & 32.23          & 79.82          & 19.74           & 7.77           & 12.24          & 61.96          & 12.24           \\
PSTNN\cite{PSTNN}     &                                      & 39.44          & 47.72          & 83.49          & 41.07           & 16.44          & 25.91          & 65.32          & 76.92           \\
IPI\cite{Gao2013InfraredPM}       &                                      & 40.48          & 50.95          & 91.74          & 148.37          & 14.40          & 31.29          & 86.35          & 450.36          \\
RIPT\cite{RIPT}      &                                      & 25.49          & 33.01          & 85.32          & 24.75           & 8.15           & 16.12          & 68.35          & 26.36           \\
NIPPS\cite{dai2016infrared}     &                                      & 33.16          & 40.91          & 80.73          & 23.64           & 16.38          & 27.10          & 70.37          & 63.27           \\ \hline
MDvsFA\cite{wang2019miss}    & \multirow{9}{*}{\rotatebox{90}{Deep learning methods}}        & 56.17          & 59.84          & 90.88          & 177.90          & 50.85          & 45.97          & 81.48          & 23.01           \\
ACM\cite{dai2021asymmetric}       &                                      & 72.45          & 72.15          & 93.52          & 12.39           & 63.38          & 60.80          & 91.58          & 15.31           \\
Res-Vit\cite{liu2021infrared}   &                                      & 72.82          & 71.22          & 98.15          & 27.15           & 61.89          & 60.64          & 90.91          & 12.64           \\
AGPC\cite{zhang2021agpcnet}      &                                      & 73.69          & 72.60          & 98.17          & 16.99           & 66.29          & 65.23          & 92.83          & 13.12           \\
DNANet\cite{li2022dense}    &                                      & 74.16          & 75.65          & 98.17          & 30.21           & 64.81          & 64.51          & 93.27          & 16.05           \\ 
HRNet\cite{sun2019deep}     &        & 76.50          & 72.86          & 99.08          & 2.88            & 64.78          & 59.47          & 92.95          & 16.95           \\
U$^2$Net\cite{U2net}  &                                      & 74.54          & 73.18          & 98.17          & 18.32           & 64.75          & 62.32          & 92.61          & 18.18           \\
SwinT\cite{liu2021swin}     &                                      & 70.53          & 69.89          & 92.19          & 33.42           & 59.89          & 58.78          & 86.59          & 17.74           \\ 
NAT\cite{hassani2022neighborhood} & & 74.33 & 71.67 & 97.25 & 10.26 & 63.23 & 62.01 & 91.92 & 15.53
\\
UCF (Ours) &                                      & \textbf{80.89}          & \textbf{78.92}          & \textbf{100.00}         & \textbf{2.26}            & \textbf{68.92}          & \textbf{69.26}          & \textbf{93.60}          & \textbf{11.01}           \\ \hline
\end{tabular}}}

    \caption{Quantitative evaluation of ISOS on SIRST and IRSTD datasets. We report pixel level metric IoU $(\%)$ and nIoU $(\%)$ and object level metric Pd ($\%$) and Fa ($10^{-6}$). All the deep learning methods outperform traditional methods and our UCF achieves the best performance in all terms of metrics on both datasets.}
    \label{total results}
\end{table*}

\begin{figure}[ht]
    \centering
    \includegraphics[width=1\linewidth]{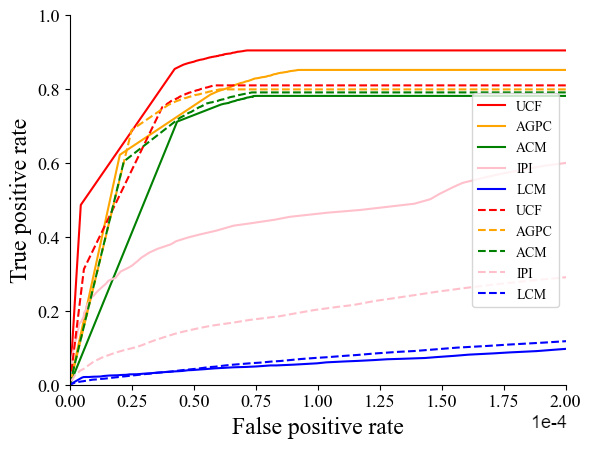}
    \caption{ROC curve on SIRST (solid line) and IRSTD (dotted line) of different methods.}
    \label{ROC}
\end{figure}

We conduct a further investigation of the dynamic relationship between Precision and Recall using LCM\cite{chen2013local}, IPI\cite{Gao2013InfraredPM}, ACM\cite{dai2021asymmetric}, AGCP\cite{zhang2021agpcnet}, and our proposed method UCF. The ROC curves are presented in \cref{ROC}, where the area under the curve (AUC) is a key metric for quantitatively evaluating the ROC. Our UCF method achieved the highest AUC and F-score on both datasets, as shown in \cref{ROC_table}.


\begin{table}[ht]
    \centering
    \caption{F-score and the area under ROC curve on both datasets with different methods.}
    \renewcommand\arraystretch{1.3}{
    \setlength{\tabcolsep}{7pt}{
    \begin{tabular}{l c c c c}
        \hline
         \multirow{2}{*}{\textbf{Method}}& \multicolumn{2}{c}{\textbf{SIRST}} & \multicolumn{2}{c}{\textbf{IRSTD}}  \\
         \cline{2-5}
          &\textbf{F-score}$\uparrow$ & \textbf{Auc}$\uparrow$ & \textbf{F-score}$\uparrow$ & \textbf{Auc}$\uparrow$\\
         \hline LCM\cite{chen2013local} & 12.80 & 0.058 & 8.52&0.099\\
         IPI\cite{Gao2013InfraredPM} & 57.63 & 0.448 & 25.17&0.248\\
         ACM\cite{dai2021asymmetric} & 84.02 & 0.684 & 77.59&0.719\\
         AGPC\cite{zhang2021agpcnet} &84.85 & 0.765 & 79.73&0.734\\
         UCF & \textbf{89.43} & \textbf{0.843} & \textbf{81.60}&\textbf{0.745}\\
         \hline
    \end{tabular}}}
    \label{ROC_table}
\end{table}

\subsection{Qualitative Comparisons}
\begin{figure*}[tp]
    \centering
    \includegraphics[width=1\linewidth]{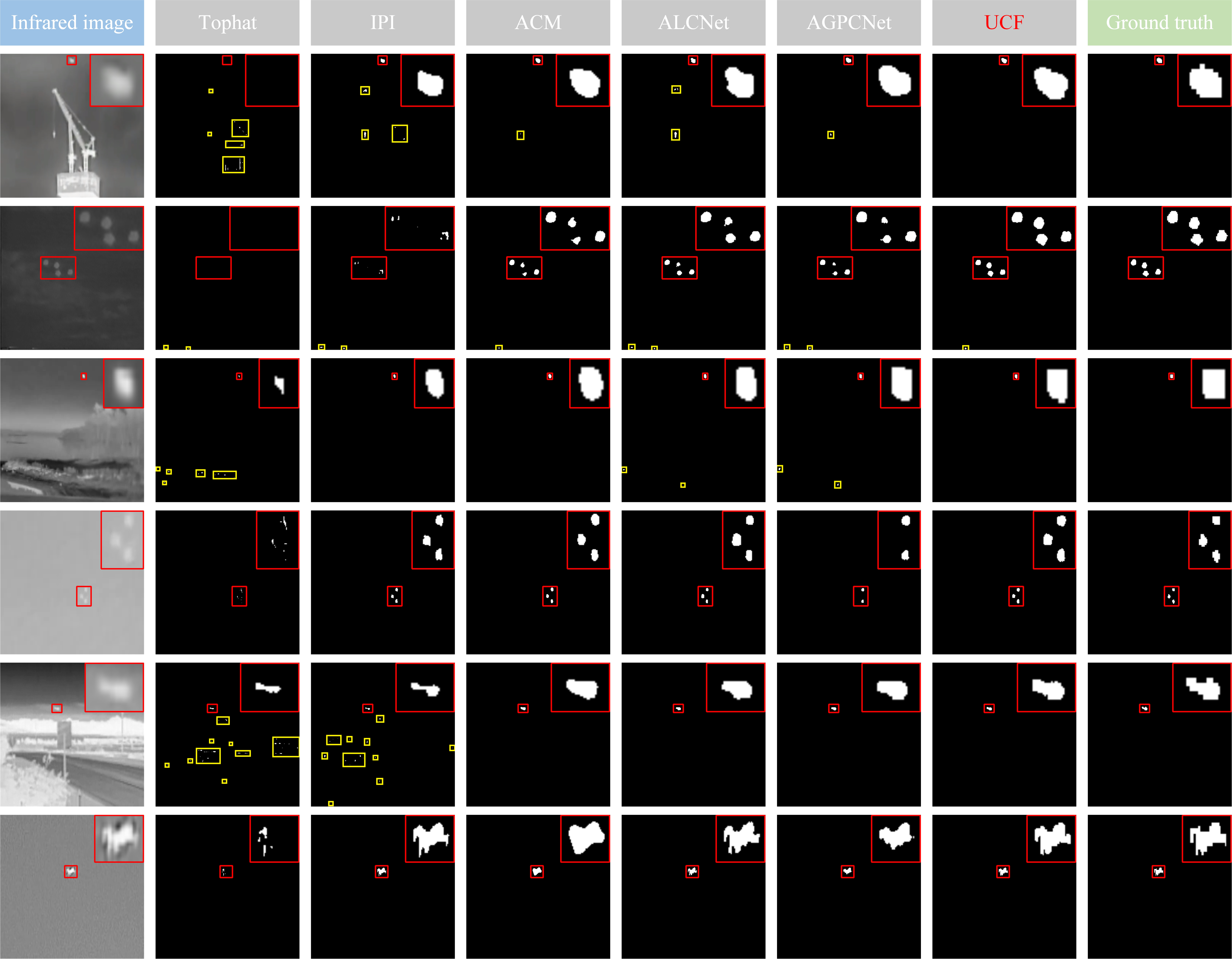}
    \caption{Qualitative results of different methods. Red bounding boxes indicate true objects and yellow bounding boxes indicate false alarms which predicted. As seen in the first four rows, in complicated backgrounds, our UCF can get less false alarm and correctly detect more targets in target level. From the last two rows, we can see that UCF describe the shape and texture in more detail in relatively simple scenes. (Best viewed by zooming in)}
    \label{qualitative_results}
\end{figure*}

Several visualization results for different methods are presented in \cref{qualitative_results}. These results clearly demonstrate that our proposed UCF method not only achieves a higher detection rate and fewer false alarms at the object level, but also predicts object shapes more accurately.

\subsection{Ablation Experiments}
We conduct a series of ablation experiments on SIRST to investigate the effectiveness of CDC and FFC. In \cref{general_result}, we demonstrate the general effectiveness of these methods. CDC improves performance by incorporating local contrast information, while FFC provides valuable global information. When combined, CDC and FFC achieve the best results in terms of IoU, nIoU, and Pd. However, the Fa score drops slightly compared to using only FFC. This is because CDC tends to fix shape and texture details, which can sometimes result in false pixels.
\begin{table}[ht]
    \centering
    \caption{Ablation study of CDC and FFC in IoU$(\%)$, nIoU$(\%)$, Pd$(\%)$, Fa$(10^{-6})$}
    \renewcommand\arraystretch{1.4}{
    \setlength{\tabcolsep}{6pt}{
    \begin{tabular}{ l l l l l }
        \hline
         \textbf{Method}& \textbf{IoU}$\uparrow$ & \textbf{nIoU}$\uparrow$& \textbf{Pd}$\uparrow$ & \textbf{Fa}$\downarrow$\\
         \hline
         UCF (vanilla)& 74.14 & 74.89 & 96.33 & 4.83\\
         UCF + CDC & 75.95 & 76.29 & 97.25 & 3.46\\
         UCF + FFC & 79.55 & 77.23 & \textbf{100.00} & \textbf{1.82}\\
         UCF + CDC + FFC & \textbf{80.89} & \textbf{78.72} & \textbf{100.00} & 2.22\\
    \hline
    \end{tabular}}}
    \label{general_result}
\end{table}

\textbf{Effectiveness of CDC.}
\label{cdc_experiments}
\begin{figure*}[t]
    \centering
    \begin{subfigure}{0.45\linewidth}
    \centering
    \includegraphics[width = 1\linewidth]{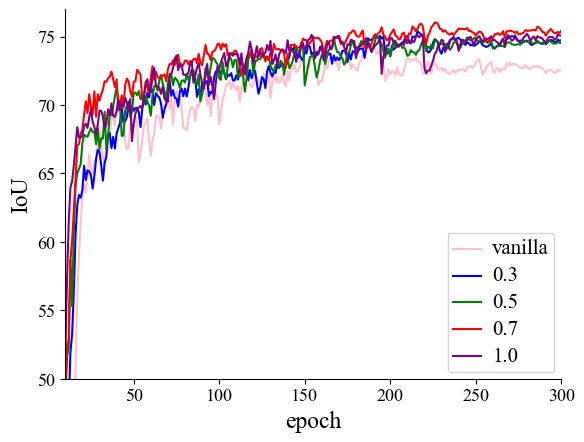}
    \caption{IoU curves on different paramater $\theta$ in CDC, the best performance is obtained when $\theta=0.7$.}
    \label{cdc_iou}
    \end{subfigure}
    \begin{subfigure}{0.45\linewidth}
    \centering
    \includegraphics[width = 1\linewidth]{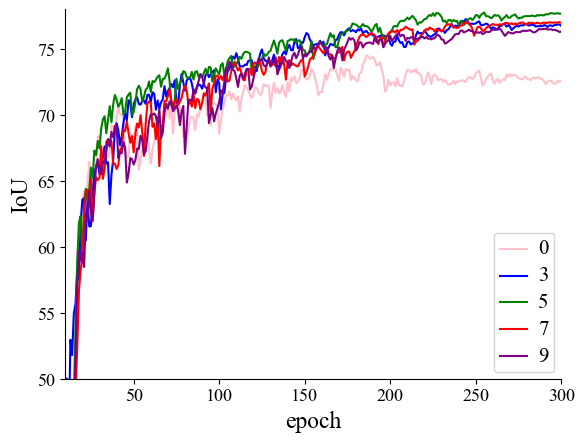}
    \caption{IoU curves using different number of FFC residual blocks, we achieve our best performance with $5$ FFC residual blocks.}
    \label{ffc_iou}
    \end{subfigure}
    \caption{IoU curves on SIRST with different parameter settings.}
    \label{total_iou_curve}
\end{figure*}
We conduct experiments with different hyperparameter $\theta$ which determines the ratio between CDC 
and vanilla convolution. As shown in \cref{cdc_iou}, CDC consistently outperforms vanilla 
covolution by mining essential contrast information and we achieve the best performance when $\theta=0.7$. 
In addtition, we compare the performance of deformable convolution and gated convolution with that of CDC. As show in \cref{cdc_ablation}, gated convolution only has a slight 
improvement over vanilla convolution, while deformable convolution has a significant drop a lot, this is because the sharp information 
is quite insufficient in ISOS tasks and deformable convolution fails to learn the offsets according to the target's shape. These results indicate that our proposed CDC has clear advantages over other convolution operators for ISOS.

\begin{table}[htp]
    \centering
    \caption{Ablation study of CDC with other convolution operators in metrics of IoU ($\%$), nIoU ($\%$), Pd ($\%$), Fa ($10^{-6})$.}
    \renewcommand\arraystretch{1.4}{
    \setlength{\tabcolsep}{6pt}{
    \begin{tabular}{ l l l l l}
    \hline
        \textbf{Conv Operator}&\textbf{IoU$\uparrow$} &\textbf{nIoU$\uparrow$} &\textbf{Pd$\uparrow$} &\textbf{Fa$\downarrow$} \\
        \hline
        Vanilla &74.14&74.89& 96.33&4.83 \\
        Gated\cite{yu2019free} & 74.31&74.57&\textbf{97.25}&23.20\\
        DeformableV1\cite{dai2017deformable} &68.00&69.60&93.58&12.60\\
        DeformableV2\cite{zhu2019deformable} &69.72&72.67 &96.33&29.90\\
        CDC ($\theta=0.7$) & \textbf{75.95} & \textbf{76.29} & \textbf{97.25} & \textbf{3.46}\\
        \hline
    \end{tabular}}}
    \label{cdc_ablation}
\end{table}
\begin{table}[htp]
    \centering
    \caption{Ablation study of FFC with other global information extracting method (vanilla conv blocks, multi-dilated conv blocks and double attention block in metrics of IoU ($\%$), nIoU ($\%$), Pd ($\%$), Fa ($10^{-6})$.}
    \renewcommand\arraystretch{1.4}{
    \setlength{\tabcolsep}{3.5pt}{
    \begin{tabular}{ l l l l l }
    \hline
        \textbf{Method}&\textbf{IoU$\uparrow$} &\textbf{nIoU$\uparrow$} &\textbf{Pd$\uparrow$} &\textbf{Fa$\downarrow$} \\
        \hline
        Vanilla Conv blocks&75.31&73.91& 98.17&12.29 \\
        Dilated Conv blocks\cite{wang2018understanding}  & 77.34&75.51&99.08&6.56\\
        Double attention\cite{fu2019dual} &77.87&75.84&\textbf{100.00}&12.51\\
        FFC blocks &\textbf{79.55}&\textbf{77.23} &\textbf{100.00}&\textbf{1.82}\\
        \hline
    \end{tabular}}}
    \label{ffc_ablation}
\end{table}
 \textbf{Effectiveness of FFC.}
 \label{ffc_experiments}
 In our FFC block, we conduct experiments to study the inner parameter of $n$ which determines the number of FFC blocks used in our method. As shown in \cref{ffc_iou}, the FFC residual block, designed to extract global context, significantly improves performance, and we achieve the best results when using five FFC residual blocks. We further explore other methods for extracting global information, such as dilated convolution and double attention, as mentioned in \cref{subsec:global_information_extraction}, from \cref{ffc_ablation} we can see that dilated convolution and double attention both show improvements in performance by enlarging receptive fields and extracting global context. However, dilated convolution only captures image-level information in deep layers, while attention mechanisms lack local inductive bias. Therefore, they are inferior to FFC in ISOS tasks.

%% file: 10_conclusion.tex
\section{Conclusion}
\label{sec:conclusion}
In this study, we identified and analyzed two important issues with ISOS that can affect model performance. Drawing inspiration from these issues, we propose a simple yet effective method. Specifically, to address the first issue of insufficient information, we use central difference convolution to guide the network's focus on local contrast information. To deal with the second issue, we employ fast Fourier convolution to extract global context from high-resolution feature maps, preventing small objects from being overwhelmed. Extensive experiments have validated that our UCF model shows great superiority over other state-of-the-art methods on both public datasets. Our proposed model can also serve as a guide for further investigations into ISOS tasks.

%% file: 12_appendix.tex
\appendix
\label{sec:appendix}
